\documentclass[letterpaper, 10 pt, conference]{ieeeconf}
\usepackage[binary-units=true]{siunitx}
\usepackage{times}
\usepackage{amsmath,amssymb}
\usepackage[nolist,nohyperlinks]{acronym}
\usepackage{accents}
\usepackage{graphicx}
\usepackage{bm}
\let\labelindent\relax
\usepackage[inline]{enumitem}
\usepackage{array}
\usepackage{multicol}
\usepackage{multirow}
\usepackage{tabulary}
\usepackage{booktabs}
\usepackage{subfig}
\usepackage{todonotes}
\usepackage{cite}
\usepackage[bookmarks=true]{hyperref}

\newcommand{\norm}[1]{\left\lVert#1\right\rVert}
\begin{acronym}
	\acro{CoM}{Center of Mass}
	\acro{UE}{User Equipment}
	\acro{UPF}{User Plane Function}
	\acro{HCOD}{Hierarchical Complete Orthogonal Decomposition}
	\acro{PR}[\emph{PR}]{purely remote control scheme}
	\acro{LA}[\emph{LA}]{locally assisted remote control scheme}
	\acro{5G}{Fifth Generation}
	\acro{QP}{Quadratic Program}
	\acro{DoF}{Degree of Freedom}
	\acrodefplural{DoF}{Degrees of Freedom}%
	\acro{leg}[\emph{L}]{\emph{legible}}		
\end{acronym}
\IEEEoverridecommandlockouts          
\renewcommand{\baselinestretch}{0.98}

\title{\LARGE \bf
Enabling Remote Whole-Body Control with 5G Edge Computing}

\author{Huaijiang Zhu$^{1}$, Manali Sharma$^{1}$, Kai Pfeiffer$^{1}$\\
Marco Mezzavilla$^{1}$, Jia Shen$^{2}$, Sundeep Rangan$^{1}$, and Ludovic Righetti$^{1, 3}$%
\thanks{$^{1}$Tandon School of Engineering, New York University, USA}%
\thanks{$^{2}$OPPO, China}%
\thanks{$^{3}$Max Planck Institute for Intelligent Systems, T\"{u}bingen, Germany}%
\thanks{Part of this work was supported by New York University, NSF grants
1936332, 1824434, 1833666, 1564142, 1925079, 1825993; NYU WIRELESS and its industrial affiliates;
NIST grant 70NANB17H166; SRC; 
and a research grant from OPPO.}%
}
\begin{document}
\maketitle
\thispagestyle{empty}
\pagestyle{empty}

\begin{abstract}
Real-world applications require light-weight, energy-efficient, fully autonomous robots.  Yet, increasing autonomy is oftentimes
synonymous with escalating computational requirements. It might thus be desirable to offload intensive computation---not only sensing and planning, but also low-level whole-body control---to remote servers in order to reduce on-board computational needs. \ac{5G} wireless cellular technology, with its low latency
and high bandwidth capabilities, has the potential to unlock cloud-based
high performance control of complex robots.
However, state-of-the-art control algorithms for legged robots can only tolerate very low control delays, which even ultra-low latency \ac{5G} edge computing can sometimes fail to achieve. 
In this work, we investigate the problem of cloud-based whole-body control of legged robots over a \ac{5G} link.
We propose a novel approach that consists of a standard optimization-based controller on the network edge and a local linear, approximately optimal controller that significantly reduces on-board computational needs while increasing robustness to delay and possible loss of communication. 
Simulation experiments on humanoid balancing and walking tasks that includes a realistic \ac{5G} communication model demonstrate significant improvement of the reliability of robot locomotion under 
jitter and delays likely to be
experienced in \ac{5G} wireless links.
\end{abstract}

\section{Introduction}
\label{sec:introduction}
Legged robots are favored in many scenarios such as disaster rescue and advanced manufacturing due to their high mobility \cite{kheddar2019}. However, to cope with dynamic, ever-changing real-world environment, robots need to be vested with fast and reliable sensing, planning and control capabilities, which all require large amount of computation. However, powerful on-board computing  increases the weight of the robot and its energy consumption, and consequently affects the autonomy of the robot. For example, ANYmal, one of the most advanced commercially available quadruped robot, carries \SI{3}{\kilo\gram} of batteries of about \SI{650}{\watt\hour} energy \cite{Hutter:2016dh} while the high-end Nvidia Titan X consumes more than \SI{250}{\watt} of power, significantly impacting battery life if
such computational power was embedded on the robot.

The idea of offloading computations to the cloud is not novel~\cite{kehoe2015survey} and has attracted a lot of attention
for tasks that do not require real-time performance.
Nevertheless, cloud computing remains elusive for latency-critical tasks
as limited bandwidth and high latency of wireless communication preclude the transmission of rich multi-modal sensor data to the cloud and the remote execution of fast millisecond scale feedback control loops.

\ac{5G} cellular network edge computing technology~\cite{hu2015mobile} in conjunction
with massive broadband links in the millimeter wave (mmWave)
bands \cite{rangan2014millimeter,ford2017achieving} offer 
unprecedented access to high bandwidth and low latency communication
that could render edge-based real-time control a reality.
In traditional wide area networks, packets are typically routed through
a centralized gateway before accessing any cloud services.
This routing can cause considerable delay---in excess of \SI{40}{\milli\second} in
current 4G LTE networks \cite{laner2012comparison}.
\ac{5G} edge computing, i.e. when the computers are placed at the edge of the network, close to the wireless base stations,
dramatically reduces the delay in the core network. At the same time,
the mmWave bands offer vast amounts of spectrum that enable
ultra-fast communication in the airlink enabling delays
of \SIrange[range-phrase=--,range-units=single]{1}{2}{\milli\second}, an order of magnitude lower delays than current 4G links
operating in traditional spectrum bands, and within the requirements of state of the art torque control methods \cite{herzog_momentum_2016}.

\begin{figure}
    \centering
    \includegraphics[width=\linewidth]{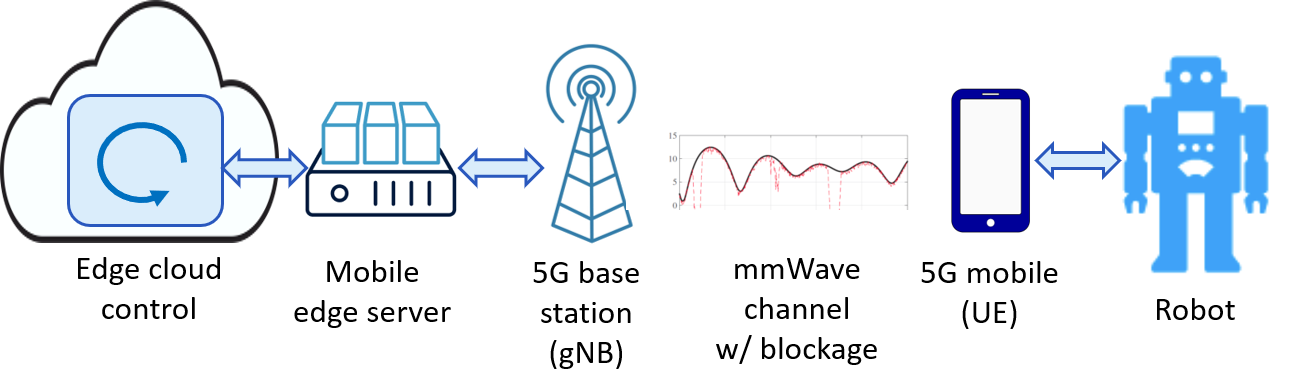}
    \caption{Cloud-based robotic 
    edge whole-body control
    over a \ac{5G} mmWave wireless link.}
    \label{fig:architecture}
    \vspace{-0.7cm}
\end{figure}

However, a key challenge with communication in the mmWave bands
is that the signals are highly susceptible to blockage
by common materials in the environment including buildings, people,
and foliage  \cite{maccartney2016millimeter,bai2013coverage,gapeyenko2016analysis,slezak2018empirical}.  Robots with metallic parts can also block mmWave signals.
As a result,
links can intermittently experience outage causing delays,
jitter and packet loss.
These outages are not permanent as signals can be re-routed or
communication can fallback to the 4G network but nevertheless
create significant delays.
Although robots may be able to detect such circumstances and adapt their behavior accordingly using models of mmWave signal propagation~\cite{channel_state_information}, such adaptation of the plan inevitably introduces a nonnegligible time window during which the robot needs to operate under increased communication delays or absence of communication with the network edge.

State-of-the-art whole-body controllers for legged robots are computationally expensive as they typically require the resolution
of quadratic programs at a control frequency of \SIrange[range-phrase=--,range-units=single]{500}{1000}{\hertz} \cite{Escande:2014en}. For example, in \cite{herzog_momentum_2016},
one core of a CPU was entirely dedicated to the computation of the
control commands.
While local computational requirements would significantly decrease
by offloading whole-body control to the edge, communication loss or delays are an important challenge as these feedback controllers are very susceptible to delays to maintain proper operating conditions and resist unexpected disturbances. This is especially important for legged robots
that need to walk in uncertain environments.

In this work, we propose and study
the first edge-based whole-body 
controller
over a \ac{5G} communication link capable of handling communication loss and delays as depicted in Fig.~\ref{fig:architecture}.
Our approach provides an efficient local control algorithm to enhance the performance of a purely remote controller in face of delays. The algorithm separates the computationally expensive part of the control problem and approximates it with a linear feedback controller based on previously obtained information from the external, full capacity controller. Together with the local information measured by the robot, it creates an approximately optimal control command to reduce the violation of task constraints and thus achieve the tasks in the presence of delays. 
We present a complete simulation environment including a realistic \ac{5G} communication model and rigid-body dynamics.
Extensive simulations demonstrate the capabilities of the approach for bipedal balancing and walking tasks in disaster response and
manufacturing environments.

\section{Background}

\subsection{Task-space inverse dynamics} 
In recent years,  optimization-based task-space inverse dynamics~\cite{saab2011generation,mansard2012dedicated,herzog2014balancing,del2015prioritized} has become ubiquitous for the control of legged robots as it provides a simple yet principled way of formulating task-space inverse dynamics problems as \acp{QP}; and more importantly, allows more flexible task descriptions using inequality constraints (e.g. to impose actuation limits or friction cone constraints).

Denoting the robot configuration as $q \in SE(3)\times \mathbb{R}^j$, the joint torque as $\tau \in \mathbb{R}^j$, where $j$ is the number of the joints, the dynamics of a robot in rigid contact can be written as
\begin{subequations}
\label{eq:eom_contact}
\begin{align}
    M\ddot{q} + h &= S^{\mathsf{T}}\tau + J^{\mathsf{T}}_cf\label{eq:eom}\\
    J_c\ddot{q} + \dot{J}_c\dot{q} &= 0\,,
\end{align}
\end{subequations}
where $M$ is the generalized inertia matrix; $h$ is  a vector of generalized forces including gravity,   centrifugal and Coriolis forces; $S^{\mathsf{T}}$ maps actuated joints torques to the generalized coordinates; $f$ denotes the contact forces, and $J_c$ is the Jacobian of the contact points.

A task function $s(q)$ maps the robot configuration to the task space, and a desired closed-loop task-space controller is given by $\ddot{s}^*$. For example, $s(t)$ can be the \ac{CoM} position; and $\ddot{s}^*$ would be the desired \ac{CoM} acceleration computed from a desired closed-loop \ac{CoM} dynamics. Task-space inverse dynamics aims to find $\ddot{q},\tau,f$ 
that satisfy physical consistency constraints while achieving several tasks as well as possible, i.e. getting $\ddot{s} - \ddot{s}^*$ as small as possible in the least-square sense. Note that $\tau$ is uniquely determined by $\ddot{q}, f$ from~\eqref{eq:eom} and that the dynamic consistency equation can be reduced to a 6D equation \cite{herzog2014balancing} by considering only the unactuated part of the dynamics. 
The problem can therefore be formulated as a \ac{QP} solving for the stacked variable $y=(\ddot{q},f)\in\mathbb{R}^n$
\begin{subequations}
\label{eq:qp_invdyn}
\begin{align}
& \underset{y=(\ddot{q},f)}{\text{minimize}} && \frac{1}{2} \sum_{i} \omega_i \norm{J_i\ddot{q}+\dot{J}_i\dot{q}-\ddot{s}_i^*}^2
\\
& \text{subject to} 
&& Ay \leq b\label{eq:constraints}
\,,
\end{align}
\end{subequations}
where $\omega_i >0$ weights the relative importance of each task and $J_i=\frac{\partial s_i}{\partial q}$ is the Jacobian of task $i$ that satisfies $\dot{s}_i = J_i\dot{q}$. Eq. \eqref{eq:constraints} summarizes all equality and inequality constraints by stacking $K$ constraints $a_k^{\mathsf{T}}y\leq b_k,\ \forall k=0,1,\dots, K$ into the matrix $A$ and the vector $b$, including the dynamic and contact constraints equations \eqref{eq:eom_contact}. Note that in this section we only describe
kinematic tasks depending on $\ddot{q}$ for simplicity but force tasks depending on $f$ can be similarly formulated and all results presented in the following trivially carry over to this case.

\subsection{Active-set method}
\label{sec:activeSet}
We solve the constrained \ac{QP}~\eqref{eq:qp_invdyn} using an active-set method. Given $y$, a constraint $a_k^{\mathsf{T}}y\leq b_k$ is called active when $a_k^{\mathsf{T}}y = b_k$. The active set $\mathcal{A}$ contains thus all equality constraints and the inequality constraints that are satisfied with equality, i.e. $a_k^{\mathsf{T}}y = b_k, \ \forall k \in \mathcal{A}$. Starting from an initial guess of the active-set, an equality constrained problem
\begin{subequations}
	\label{eq:qp_invdyn_active}
	\begin{align}
	& \underset{y=(\ddot{q},f)}{\text{minimize}} && \frac{1}{2} \sum_{i} \omega_i \norm{J_i\ddot{q}+\dot{J}_i\dot{q}-\ddot{s}_i^*}^2
	\\
	& \text{subject to} 
	&& \bar{A}y = \bar{b}\label{eq:active_set}
	\,,
	\end{align}
\end{subequations}
is solved at each iteration of the active-set search. $\bar{A}$ and $\bar{b}$ represent the constraints in the current iterate of the active set. If at the current solution ${y}^*$ to ~\eqref{eq:qp_invdyn_active}, there are inequality constraints being violated, one of them will be added (activated) to the active set. Otherwise, an inequality constraint that prevents the solution from going closer to the optimum will be removed (deactivated). The algorithm converges if no inactive violated constraints remain and no active constraints need to be deactivated. At this point the optimal active-set $\mathcal{A}_{\text{opt}}$ is found, and we denote by $\bar{A}_{\text{opt}}, \bar{b}_{\text{opt}}$ the stacked matrix and vector of the constraints $a_k^{\mathsf{T}}y = b_k, \ \forall k \in \mathcal{A}_{\text{opt}}$.

The solution ${y}^*$ to~\eqref{eq:qp_invdyn_active} is found by the Nullspace method~\cite{Nocedal2006}. A structured formulation is given by the publicly available solver in~\cite{Escande:2014en}: here the solution  
\begin{equation}
\label{eq:qp_sol}
y^* = \underline{A}^{\ddagger}\underline{b}
\end{equation}
is found with the hierarchical inverse $\underline{A}^{\ddagger}$. 
$\underline{A}$ and $\underline{b}$ denote the stacked matrix and the stacked vector
\begin{align}
\underline{A} &=
\begin{bmatrix}
\bar{A}^\mathsf{T} & w_1J_1^\mathsf{T} & \dots & \end{bmatrix}^\mathsf{T}\\
\underline{b} &=
\begin{bmatrix}
\bar{b}^\mathsf{T} & w_1(-\dot{J}_1\dot{q}+\ddot{s}_1^*)^\mathsf{T} & \dots & \end{bmatrix}^\mathsf{T}\,,
\end{align} 
respectively. Note that $\underline{A}^{\ddagger}$ is only given implicitly when computing $y^*$ by a forward recursion using the \ac{HCOD} of $\underline{A}$. This is implied throughout the paper when using the expression $\underline{A}^{\ddagger}$. 
The computation of the \ac{HCOD} requires $O(2n^3)$ operations, while the complexity of the forward recursion is $O(n^2)$ with $n$ being the number of decision variables. This scheme is potentially applicable to hierarchical problems with any number of priority levels. 

In the following, $\underline{A}^{\ddagger}_{\text{opt}}$ is associated to the \ac{HCOD} of the optimal active set $\bar{A}_{\text{opt}}, \bar{b}_{\text{opt}}$ found for~\eqref{eq:qp_invdyn_active}.

\subsection{\ac{5G} Cloud 
Edge Computing}
Our goal is to study robotic control in scenarios where
some of the computation is offloaded to an edge server over a \ac{5G} mmWave
wireless link.
The basic model is shown in Fig.~\ref{fig:architecture}.
The robot is equipped with a
wireless \ac{5G} mobile device,
called the \ac{UE},
which is functionally similar to a smartphone.
The \ac{UE} communicates wirelessly to
a base station over a mmWave channel.
In \ac{5G} terminology, the base station
is called the gNB \cite{parkvall2017nr,3GPP38.300}.
The mmWave bands are a key component 
of the \ac{5G} standard which use high bandwidth signals transmitted in 
narrow electronically steerable beams.
These signals offer massive peak
rates ($>\SI{1}{\giga\bit / \second}$) 
with very low latency (\SIrange[range-phrase=--,range-units=single]{1}{2}{\milli\second}) over the airlink.  

The latency over the airlink (the wireless connection between the \ac{UE} and gNB)
is not the only component
of delay.  In a traditional 4G cellular network, data must be typically routed to centralized gateway before it can access any third-party cloud services~\cite{s1,pdn}. 
This architecture can add considerable delay---often in excess of \SI{40}{\milli\second}
\cite{laner2012comparison}.
To enable
low end-to-end delay, \ac{5G} networks can combine a low latency airlink with
mobile edge cloud architecture~\cite{hu2015mobile}.  As shown in Fig.~\ref{fig:architecture},
data from the base stations can be routed to a mobile edge server that can host 
edge cloud services that have much lower delay to the base stations---potentially as low
as a few milliseconds, depending on the deployment. 
As per the latest \ac{5G} system architecture~\cite{mec} defined by 3GPP the \ac{5G} Core Network selects a \ac{UPF} close to the \ac{UE} and executes the traffic steering from the \ac{UPF} to the local Data Network via a N6 interface.

The basic problem we consider
in this work is how to partition the control between the local computation on the robot and remote computations on the edge server.  The key challenge is that, while \ac{5G} mmWave 
links offer very high peak rates, the signals are intermittent due to blockage
as discussed in the introduction
\cite{maccartney2016millimeter,bai2013coverage,gapeyenko2016analysis,slezak2018empirical}.
Thus, we wish to find distributed control policies that can exploit low-latency cloud resources when the wireless links are available,
but are robust during blockage and outage events.

\section{ 
Locally assisted remote whole-body
control}
\label{sec:methods}
\subsection{Structure of the \ac{QP} solution}
The \ac{QP} solution~\eqref{eq:qp_sol} has several interesting properties: First, the matrix $\underline{A}^\ddagger_{\text{opt}}$ depends only on the task Jacobians $J_i$ and the optimal active set normal $\bar{A}_{\text{opt}}$.
The matrix $\bar{A}_{\text{opt}}$ will change as a function of $q$ and $\dot{q}$, but this change---similarly to the Jacobians---will be rather slow. Thus, the matrix $\underline{A}^{\ddagger}_{\text{opt}}$ will not change too much in a short period of time as long as the optimal active set $\mathcal{A}_{\text{opt}}$ remains the same. This implies that a delayed $\underline{A}^{\ddagger}_{\text{opt}}$ will still approximately enforce all the active constraints if $\underline{b}_{\text{opt}}$ is updated sufficiently fast. 
Fortunately, $\underline{b}_{\text{opt}}$ only depends on the generalized forces $h$, the robot state $q,\dot{q}$, the task reference $\ddot{s}^*$, the task Jacobian $J_i$, and its time derivative $\dot{J}_i$. These quantities can all be stored and updated efficiently without querying a remote server, as forward kinematics and Jacobian computations can be done in $O(n)$, where $n$ is the robot's number of \acp{DoF}. Constructing $\underline{b}_{\text{opt}}$ from these quantities also only requires basic matrix operations that a low-power on-board computer can easily execute. Notably, all the error feedback terms are contained in $\underline{b}_{\text{opt}}$ which therefore renders it the most important quantity for control, while the matrix $\underline{A}^{\ddagger}_{\text{opt}}$ can
be seen as a projector that changes slowly.

These properties naturally partition the optimal control command into two parts: 
\begin{enumerate}
    \item the computation of $\underline{A}^{\ddagger}_{\text{opt}}$ which is expensive but less susceptible to delays; and
    \item the construction of $\underline{b}_{\text{opt}}$ which contains the error feedback terms and is latency-critical but can be done efficiently.
\end{enumerate}

In this paper, we propose to offload the active-set search and the computation of $\underline{A}^{\ddagger}_{\text{opt}}$ to the remote server, while updating $\underline{b}_{\text{opt}}$ locally on the robot. If the robot fails to recover the latest decomposition $\underline{A}^{\ddagger}_{\text{opt}}$ due to communication delays or packet loss, we approximate it with the most recently received one. This still enables us to perform feedback control for all the tasks and to approximately enforce previously active inequality constraints.

\subsection{Control scheme}
\begin{figure}[!t]
    \centering
    \includegraphics[width=0.9\linewidth]{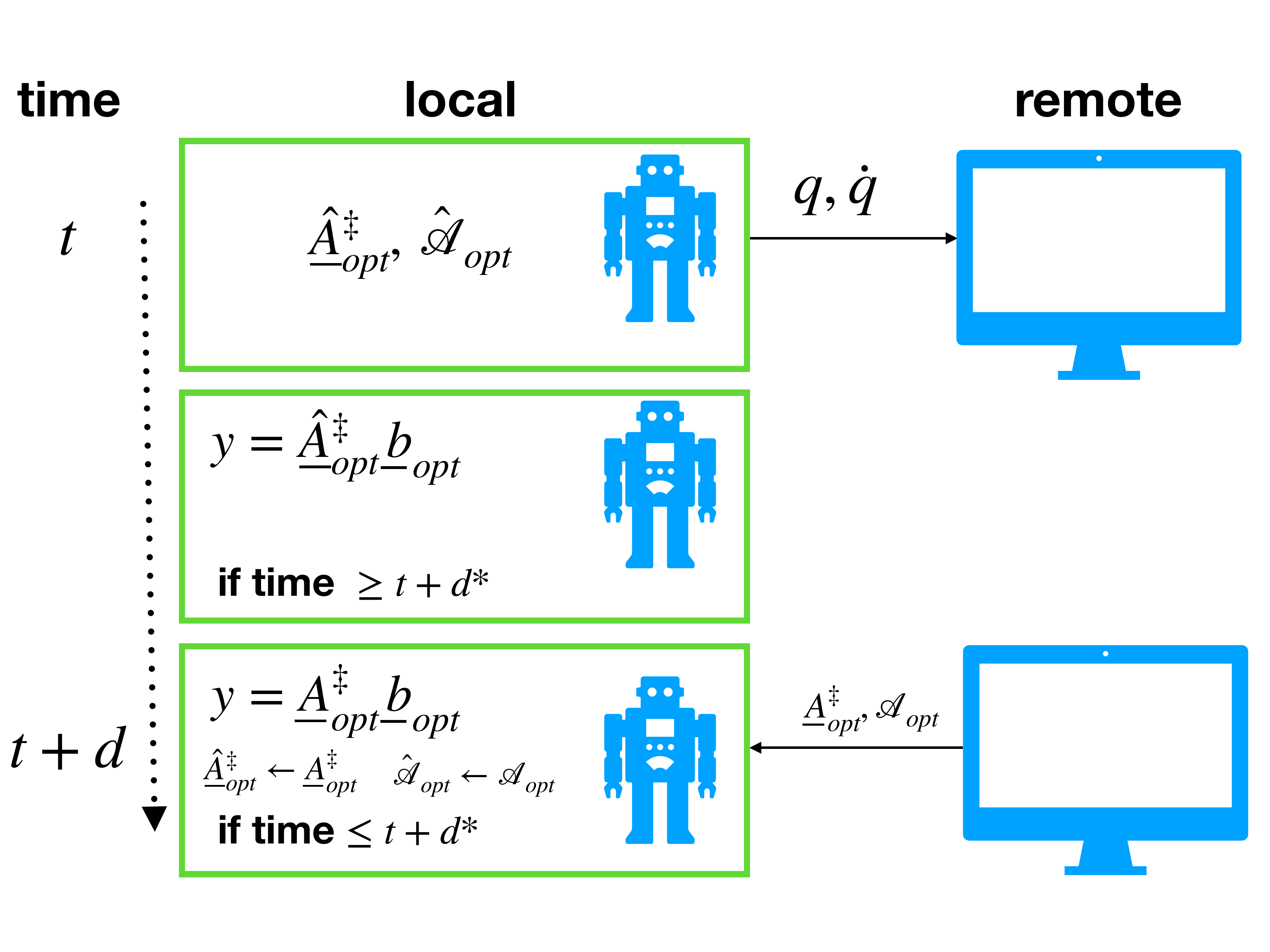}
    \caption{Communication and computation flow between the robot and the edge computer in the proposed control scheme.}
    \label{fig:scheme}
    \vspace{-0.6cm}
\end{figure}

These insights together give rise to the following control scheme illustrated in Fig.~\ref{fig:scheme}. At any time $t$, the robot maintains a cache of the decomposition $\underline{\hat{A}}^{\ddagger}_{\text{opt}}$ and the optimal active set $\mathcal{\hat{A}}_{\text{opt}}$ from the most recently successful communication with the remote server, and the robot  measures its state $q,\dot{q}$ and constructs the vector $\underline{b}_{\text{opt}}$; then it sends $q,\dot{q}$ to the remote controller to solve the full \ac{QP} problem~\eqref{eq:qp_invdyn}. Meanwhile, the robot computes the approximate command 
\begin{align*}
    \hat{y} = \underline{\hat{A}}^{\ddagger}_{\text{opt}}\underline{b}_{\text{opt}}\,.
\end{align*}
 Note that the active-set search is not continued according to the changed right hand side $\underline{b}_{\text{opt}}$ as it is potentially expensive if a lot of active-set iterations are necessary. Instead, the last found optimal active-set $\mathcal{\hat{A}}_{\text{opt}}$ is used when computing the approximate solution.

At time $t+d$, the robot receives a new decomposition $\underline{{A}}^{\ddagger}_{\text{opt}}$ and the new optimal active set $\mathcal{A}_{\text{opt}}$ from the remote controller, where the delay $d$ depends on the communication channel. The local cache is updated accordingly. Note that this solution may not have been computed based on the state $q,\dot{q}$ measured at time $t$ due to the previous delays. 

Given the period of the control loop $T$ (e.g. $T=\SI{1}{\milli\second}$), we can set a desired threshold $0<d^*<T$ such that the robot applies a control command
\begin{align}
    y = 
    \begin{cases}
     \underline{A}^{\ddagger}_{\text{opt}}\underline{b}_{\text{opt}} &\text{ if } d < d^*\\
    \hat{y} &\text{ otherwise }\,.
    \end{cases}
\end{align}
At time $t+T$, the process described above repeats.

\subsection{Handling contact switches}
\label{sec:onboard}
The control scheme described above relies on the assumption that the optimal active set does not change when the communication delay occurs. This assumption is plausible for tasks where no contact is broken or established when blocking events occur, such as balancing without external disturbances. However, for contact-switching tasks such as walking, this assumption becomes problematic for two reasons
\begin{enumerate*}
    \item the contact switches introduce very different constraints and thus different optimal active sets; and 
    \item even in the same contact mode, when the robot approaches the contact switch, the optimal active set changes more frequently due to the friction cone constraints being activated.
\end{enumerate*}

To resolve the first issue, we pre-compute the solution to the new contact mode \textit{shortly before} the contact switch by imposing the new contact constraints on the current robot state. This gives a valid approximation for low-speed locomotion because the state of the robot $q, \dot{q}$ shortly before and after the contact switch is similar. The solution differs mostly due to the different contact forces and the corresponding constraints. Therefore, if the contact switch does happen and we have not yet obtained the corresponding optimal solution from the remote server, we can use the pre-computed solution and approximately enforce the new constraints. On the other hand, for highly dynamic locomotion, the generalized velocity $\dot{q}$ may change significantly across contact switches; to tackle this issue, a more sophisticated prediction of the state $q,\dot{q}$ is required, which we leave for future work.

While this pre-computation can in principle be performed on the 
edge computer, it has to be completed before the contact switch occurs. However, it is difficult to have a delay upper bound in the \ac{5G} network and this upper bound might anyway be too large compared to the timing of one step. In this case, we additionally use the on-board computer to perform the pre-computation. As we assume that the on-board
computer has very limited computational capabilities, the \ac{QP}
needs to be started several control cycles prior to the switch.
The delay introduced by the on-board computation can be considered upper-bounded by a constant in practice as will be shown in the simulation experiments.

The second issue can also be mitigated in a similar manner---we can solve the full \ac{QP} problem~\eqref{eq:qp_invdyn} on-board as well when the contact switch is happening, as the optimal active set is more likely to be similar in a shorter period.

In our implementation, in the time window of length \SI{100}{\milli\second} centered at the planned contact switch time, we perform all computation---including the full \ac{QP}, the local controller, and the pre-computation (once per contact switch) of the next contact mode---locally, i.e. on the on-board computer. Note, in this case we choose to solve the full \ac{QP} every \SI{5}{\milli\second} due to limited computational capacity. In addition, the pre-computation will be initiated \SI{10}{\milli\second} before the planned contact switch time. The aforementioned choice of numbers is reasonable for our simulated low-power on-board computer, as later simulation experiments will show that it takes less than \SI{3}{\milli\second} to solve the full \ac{QP} locally in the worst case.
We do still query the remote server to solve the full \ac{QP} at the same time, so that we get better approximation when the communication with the server is faster than the on-board computation. The increase in on-board computational complexity is minimal as full \acp{QP} are only solved on-board in a short period of time around contact switches at a much lower speed than the control frequency.

It is worth noting that the way we handle contact switches as described above heavily relies on the accurate knowledge of when and how the contact switch will happen, resulting in lack of robustness of our approach to unexpected contacts. However, a principled handling of unexpected contact switches is still an open problem for optimization-based task-space inverse dynamics controllers even without control delays. Addressing this issue
therefore goes beyond the scope of this paper and we leave the question 
of robustness to unexpected contact changes to future research.

\section{Simulation experiments}
In the following simulation experiments, we compare our \ac{LA} with a \ac{PR}. \ac{PR} sends the measured robot state to the remote controller to compute the optimal solution. If the robot does not recover the latest command from the remote controller,  executes the most recently received command. The goal of these experiments is to demonstrate that
our approach significantly improves robustness to delays with limited
computational overhead.

The simulation experiments consist of robot balancing and walking tasks under two different delay settings: constant delays and simulated stochastic delays in \ac{5G} networks. The simulations were conducted on an Intel Xeon CPU at \SI{3.7}{\giga\hertz}. The on-board computer of the robot is emulated by restricting the computation to a single core of the CPU at \SI{1.2}{\giga\hertz}. We simulate the $37$-\ac{DoF} humanoid robot Romeo and use Pinocchio~\cite{pinocchioweb} for rigid body dynamics computation.

\label{sec:experiments}
\subsection{Robot tasks}
Our control scheme is evaluated on two typical tasks for legged robots, namely balancing and walking.
\begin{enumerate}
    \item \textbf{Balancing}
    The task is achieved by stabilizing the \ac{CoM} of the robot while maintaining rigid contacts between the feet and the ground. To prevent slipping, we constrain the ground reaction forces to stay inside the friction cones approximated with $4$-sided pyramids. In addition, we minimize an error between the current joint positions and a desired upright posture in the whole-body joint space to resolve remaining torque redundancies. We inject Gaussian noise with a noise level $\sigma = 1\times 10^{-2}$ to the joint measurements and apply an external push of \SI{100}{\newton} for \SI{0.2}{\second} at the robot base.
    \item \textbf{Walking} For this task, we first generate offline a desired \ac{CoM} trajectory and a footstep plan using a linear inverted pendulum model~\cite{herdt2010online}. The motion of the swing foot is synthesized by interpolating splines from one step location to another and each step takes \SI{0.8}{\second}. The robot alternates between single support phase (one foot on the ground) and short double support phases of \SI{5}{\milli\second} (both feet on the ground) by following the desired \ac{CoM} trajectory and foot motion. In the double support phase, both feet have to obey rigid contact constraints and friction cone constraints, while in the single support phase, only the support foot needs to respect those two constraints. The measurement noise level is $\sigma = 1\times 10^{-3}$ in this task.
\end{enumerate}
 The task objectives and constraints are formulated by TSID~\cite{tsidweb} and then solved as a constrained \ac{QP} by the solver~\cite{Escande:2014en} as described in Sec.~\ref{sec:activeSet}.


\subsection{Network and wireless modeling}
\label{sec:network_sim}
The communication between the server and the robot over the \ac{5G} channel is simulated using ns-3~\cite{henderson2008network}, a widely-used open-source network simulator.
We use a state-of-the-art \ac{5G} mmWave module in ns-3 to simulate the wireless link and network stack
\cite{mezzavilla2018end} used in several works
\cite{zhang2019will,ford2016framework}.
This module includes detailed models for the wireless channel, 
wireless communication stack, core network and networking protocols.
We use \verb|MmWave3GPPPropogationLossModel| to configure the communication channel. Note that at the time of writing, real \ac{5G} base stations were 
not available.

\begin{figure}[tp]
	\centering
	\subfloat[No blockage\label{fig:no_blockage}]{\includegraphics[width=0.5\linewidth]{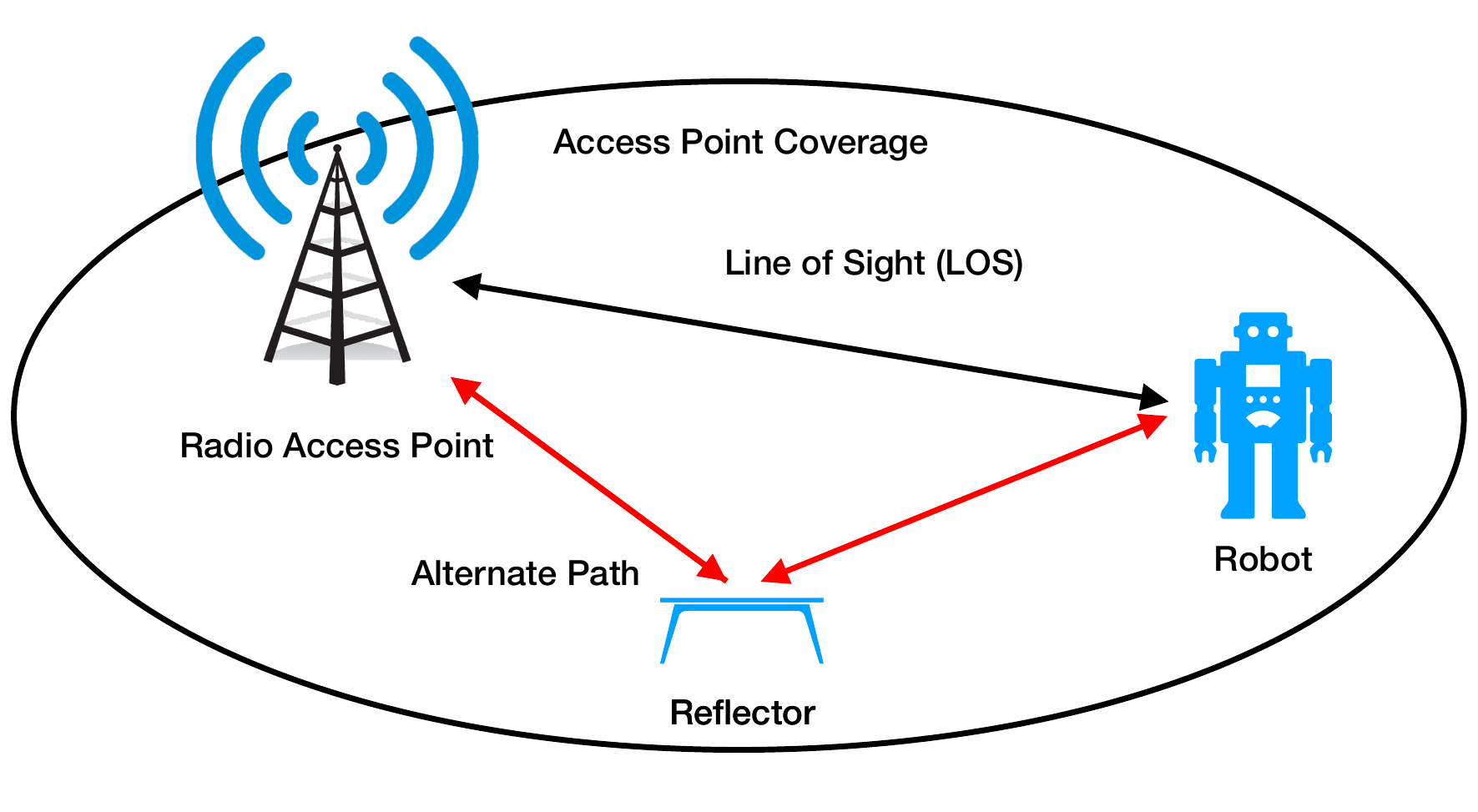}}
	\subfloat[With blockage\label{fig:with_blockage}]{\includegraphics[width=0.5\linewidth]{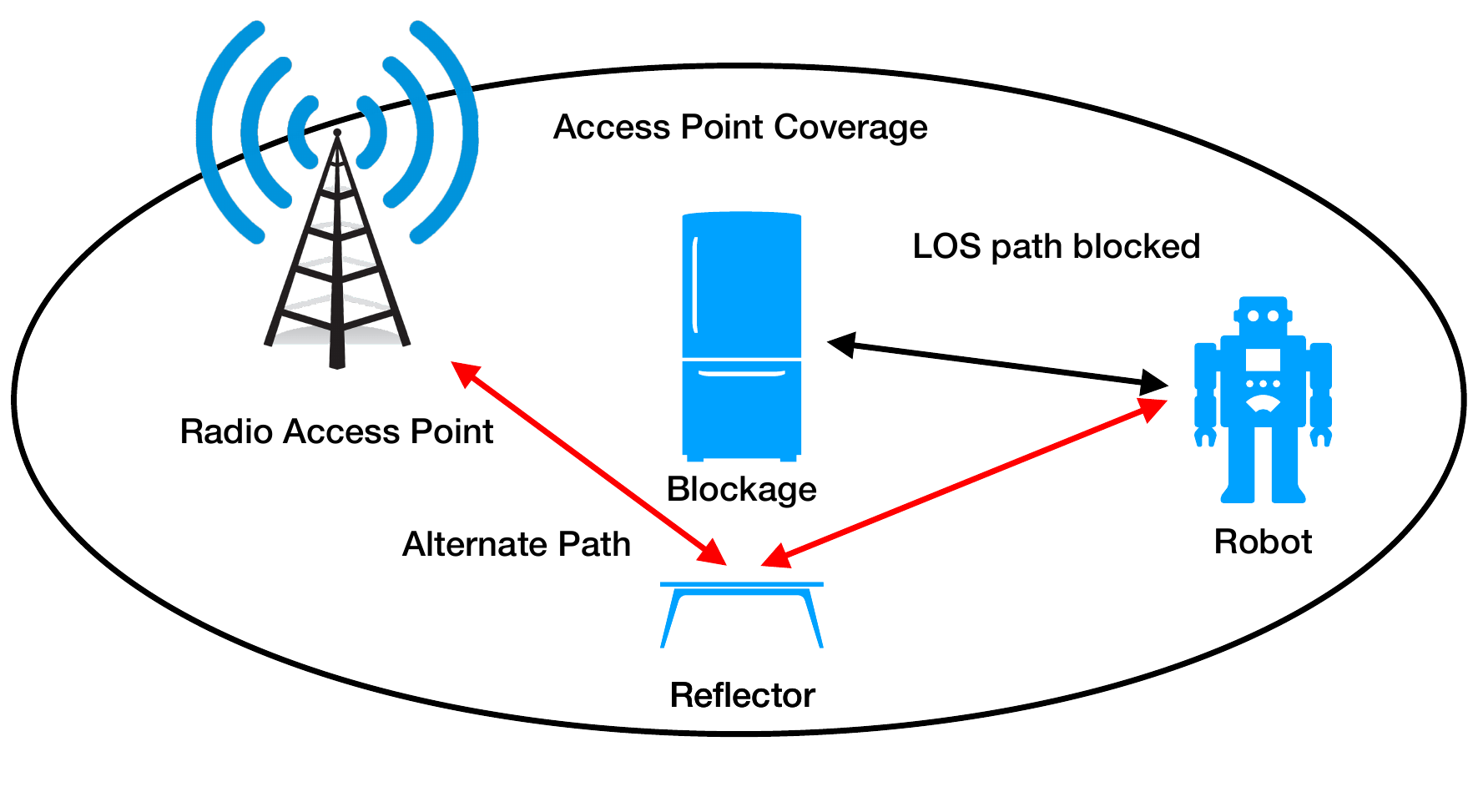}}
	\caption{Illustration of Line of Sight and Multipath of the signal for non-blocking and blocking scenarios respectively}
	\label{fig:blockage}
	\vspace{-0.6cm}
\end{figure}

Our key task is to 
analyze the robustness due
to blockage.
To model this, we follow
\cite{ford2016framework,zhang2019will} where
several moving obstacles are placed in the environment
and blockage is computed via ray tracing.  We assume fixed number of blockers in the
environment and fully digital beamforming to expedite beam search during
blocking events.

Fig.~\ref{fig:with_blockage} illustrates the blocking scenario in the simulation. 
The blockages move at a speed of \SI{2}{\meter/\second}. 
Table~\ref{tab:ns3} details the parameters used in the ns-3 simulation. The type of blockages and the distance between the access point and the robot is dependent on the simulation scenarios.
We consider two
scenarios relevant for legged robot applications (parameters for
both scenarios can be found in
\cite{mezzavilla2018end}):
\begin{enumerate}
    \item \textbf{Smart factory} This scenario models the case where robots conduct manufacturing tasks autonomously or with humans. The access point is placed in the middle of the factory; the robot or the blockages can move around it. Solid metal and stone blockages are used in this scenario. It is characterized by infrequent blocking events as the factory environment is well structured. However, delays introduced by blocking events can be as high as \SI{389}{\milli\second} as potential blockages such as containers and manufacturing components are large in size.
    \item \textbf{Burning building} This is a mission critical scenario, where a robot is sent inside a burning building to analyze and report back possible threats. The access point is placed on a window while the robot is mobile inside. The blockages in this case are in close proximity of each other and are smaller in size. The type of blockages used are wooden or solid metal. This results in a delay profile characterized by frequent but lower delays; the maximal delay in this scenario does not exceed \SI{91}{\milli\second}. \end{enumerate}
    We also note that the simulated delays do not reflect the possible damage to the access point in the burning building scenario.

\begin{figure}[tp]
	\centering
	\subfloat[Smart factory\label{fig:delay_factory}]{\includegraphics[width=0.5\linewidth]{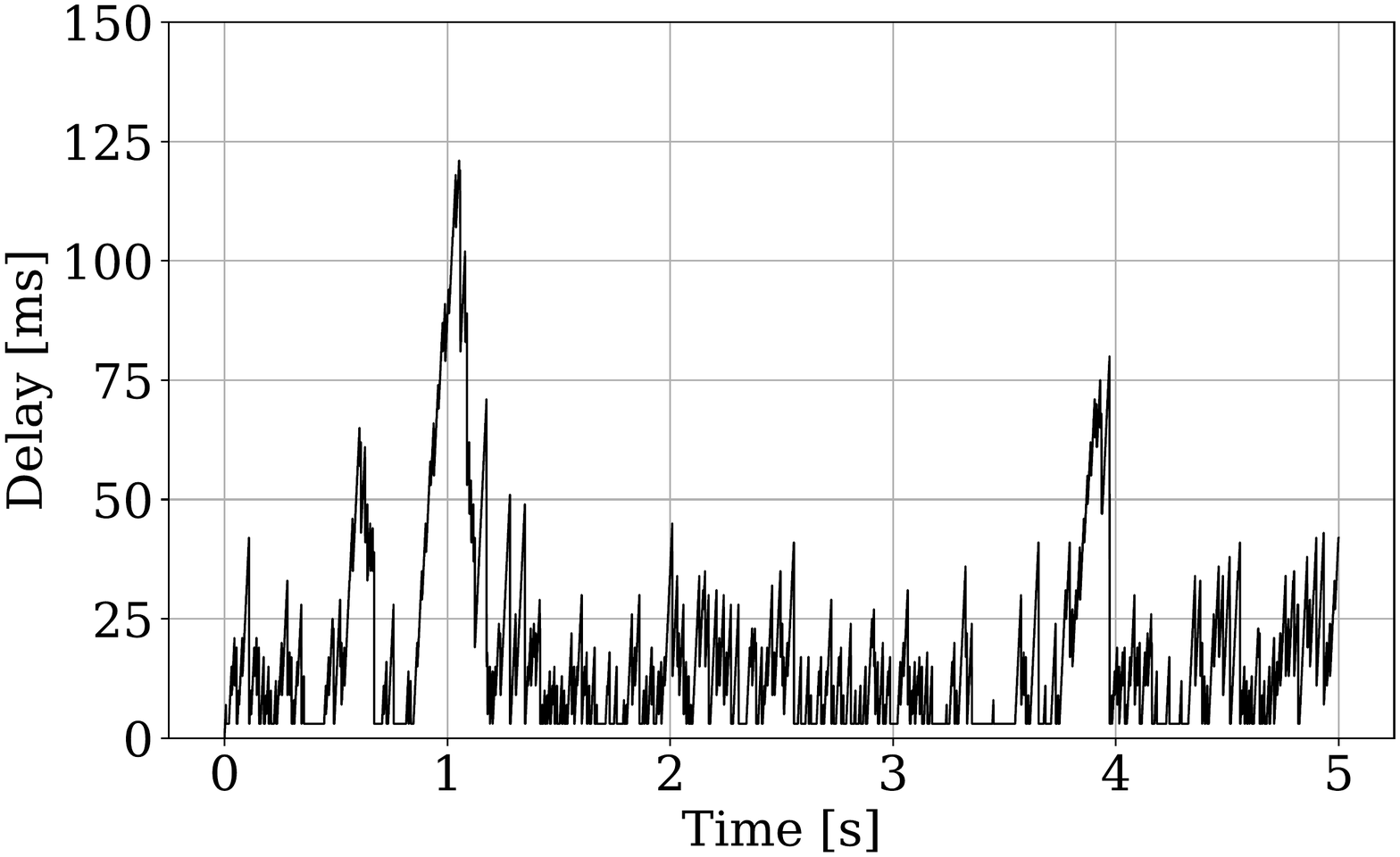}}
	\subfloat[Burning building\label{fig:delay_building}]{\includegraphics[width=0.5\linewidth]{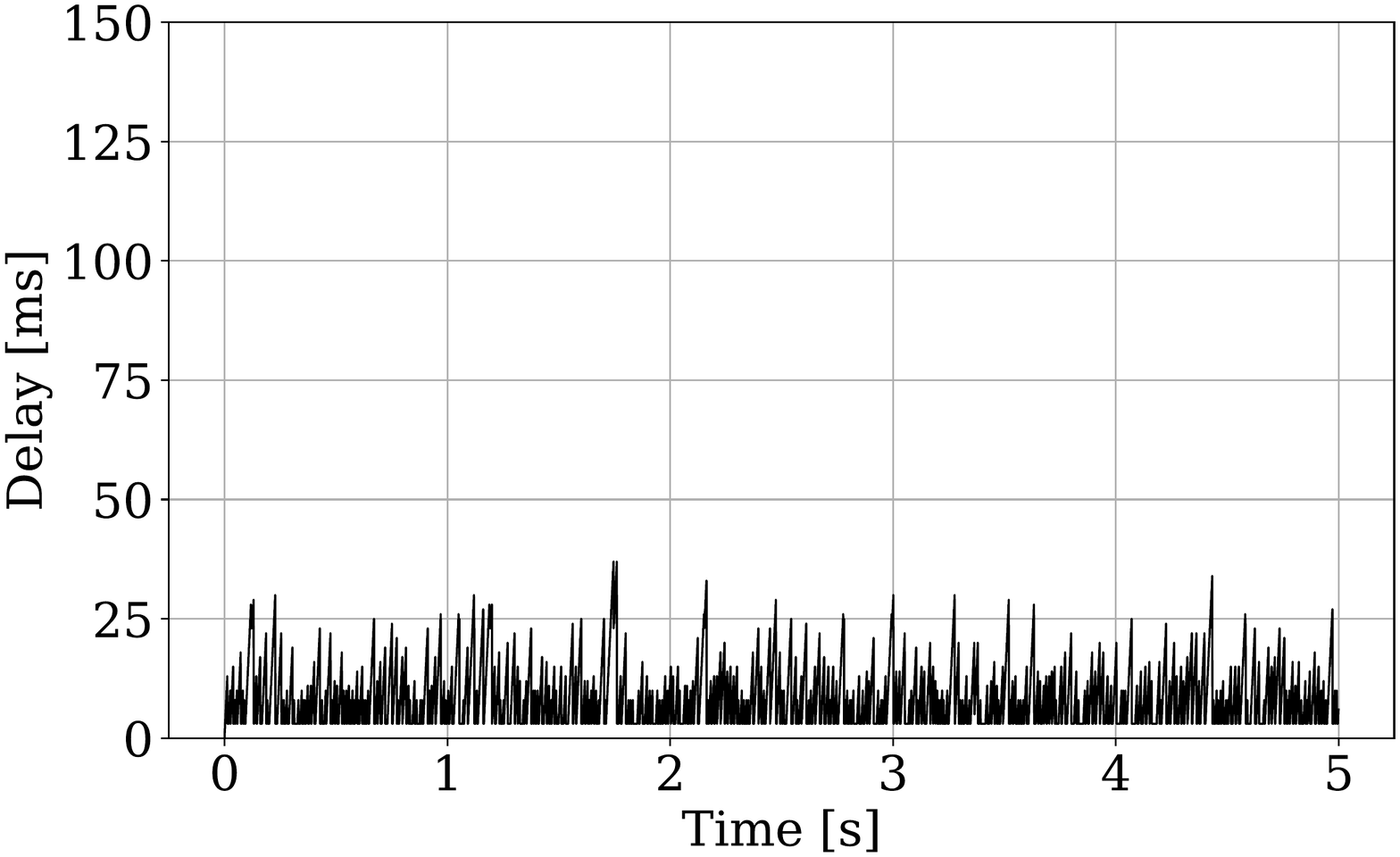}}
	\caption{An instance of the delay profile in each of the simulated scenarios.}
	\label{fig:simulated_delay}%
	\vspace{-0.3cm}
\end{figure}

\begin{table}[tp]
\centering
\caption{ns-3 simulation parameters}
\label{tab:ns3}
\begin{tabular}{lcc}
\toprule
\multicolumn{1}{c}{\textbf{\scriptsize Parameters}} & \textbf{\scriptsize Smart factory} & \textbf{ \scriptsize Burning building} \\
\midrule
\scriptsize Channel model                    & \scriptsize UMi-StreetCanyon       & \scriptsize InH-OfficeMixed           \\
\scriptsize Maximal distance between                & \multirow{2}{*}{\SI{100}{\meter}} & \multirow{2}{*}{\SI{30}{\meter}}     \\
\scriptsize the access point and robot              &                        &                           \\
\scriptsize Bandwidth                               & \multicolumn{2}{c}{\SI{1}{\giga\hertz}}                           \\
\scriptsize Number of blockers                      & \multicolumn{2}{c}{4}                              \\
\scriptsize Transport layer protocol                & \multicolumn{2}{c}{UDP}                            \\
\scriptsize Maximal uplink packet size              & \multicolumn{2}{c}{\SI{1}{\kilo\byte}}                           \\
\scriptsize Maximal downlink packet size            & \multicolumn{2}{c}{\SI{40}{\kilo\byte}}                          \\
\scriptsize Transmission frequency                 & \multicolumn{2}{c}{\SI{1}{\kilo\hertz}}\\
\bottomrule
\end{tabular}
\vspace{-0.6cm}
\end{table}

In both scenarios above, we have assumed
a delay from the base station to the mobile edge
server of
\SI{1}{\milli\second}, a realistic value
for future \ac{5G} edge deployments.
At the transport layer, we have assumed UDP
instead of TCP.
TCP ensures packet delivery using re-transmissions 
whereas UDP transmits only once and does not wait for any acknowledgement or ensure packet delivery. 
Since our robotics automation loop discards any packet that does not meet the specified time constraints, re-transmission would only lead to excessive bandwidth consumption and cause more delay.

Fig.~\ref{fig:simulated_delay} shows an instance of the delay profile in each of the scenarios under
these assumptions. We generated $100$ different delay profiles for each scenario by randomizing the initial positions of the robot and the blockages. The delay profiles are then introduced into the robotics simulator to assess the effect of the delay.

\subsection{Metrics}
We examine the control performance by measuring 
the average \ac{CoM} tracking error and the average violation of the rigid contact constraint
\begin{align*}
    \frac{1}{N}\sum_{n=0}^N\norm{p-p_d}\, \text{ and }
    \frac{1}{N}\sum_{n=0}^N\norm{J_c\ddot{q}+\dot{J}_c\dot{q}}\,,
\end{align*}
where $N$ is the total number of the simulation steps; $p$ and $p_d$ denotes the actual and desired \ac{CoM} position; the time indices are dropped for notational simplicity. Due to symmetry, we only report the constraint violation of the left foot for the balancing task; for the walking task, we report the constraint violation of the support foot.
 
Finally, whether a robot falls or not is used as a qualitative metric to determine the failure or success of a task execution.

\subsection{Results}

\paragraph{Constant delays}
Table~\ref{tab:const_balance} and Table~\ref{tab:const_walk} report the performance of \ac{PR} and \ac{LA} in the balancing task and the walking task respectively. The infinity symbol $\infty$ in the tables indicates that the robot fell. The last row of the tables shows the maximal tolerable delay for \ac{LA} to achieve the task without the robot falling. Recall that we require on-board computation for the walking task---here we simply assume that the on-board computation causes the same respective constant delays; for instance, a \SI{25}{\milli\second} constant delay means that we solve the full \acp{QP} onboard every \SI{25}{\milli\second}. Different constant delays can be interpreted as the usage of different on-board computational capacities. Across all delay levels, \ac{LA} had lower tracking error and lower constraint violation than \ac{PR}. The maximal tolerable constant delay was significantly increased by incorporating the local controller in both the balancing task and the walking task. An important implication of this result is that our approach can also be used to address the delay caused by limited on-board computational resources if the on-board computation scheme is properly scheduled to produce bounded delays, enabling purely local optimization-based whole-body control on a low-power on-board computer. It is thus interesting for future research to investigate the control performance and power consumption of such schemes compared to our locally assisted remote control scheme.
 
\begin{table}[!t]
\centering
\caption{Balancing task performance of \ac{PR} and \ac{LA} under various constant delays. The last row shows the maximal delay \ac{LA} can tolerate.}
\label{tab:const_balance}
\begin{tabular}{ccccc}
\toprule
\multirow{2}{*}{\textbf{Delays}} & \multicolumn{2}{c}{\textbf{\ac{CoM} error }[\SI{}{\centi\meter}]} & \multicolumn{2}{c}{\textbf{Constraint violation }[\SI{}{\meter\per\second^2}]}  \\
& \ac{PR}     & \ac{LA}     & \ac{PR} & \ac{LA}    \\
\midrule
\SI{0}{\milli\second}    & 1.30            & 1.30      & 0.00          & 0.00     \\
\SI{10}{\milli\second}    & 1.35            & \textbf{1.34}      & 2.18           & \textbf{0.04}     \\
\SI{20}{\milli\second}    & 1.43            & \textbf{1.38}      & 2.17           & \textbf{0.04}     \\
\SI{30}{\milli\second}    & 1.63            & \textbf{1.46}      & 2.21           & \textbf{0.05}     \\
\SI{40}{\milli\second}            & 2.21            & \textbf{1.57}      & 2.40           & \textbf{0.05}     \\
\SI{50}{\milli\second}            & $\infty$           & \textbf{1.73}      & $\infty$           & \textbf{0.06} \\
\SI{90}{\milli\second}            & $\infty$           & \textbf{2.53}      & $\infty$           & \textbf{0.16} \\
\bottomrule
\end{tabular}
\end{table}

\begin{table}[!t]
\centering
\caption{Walking task performance of \ac{PR} and \ac{LA} under various constant delays. The last row shows the maximal delay \ac{LA} can tolerate.}
\label{tab:const_walk}
\begin{tabular}{ccccc}
\toprule
\multirow{2}{*}{\textbf{Delays}} & \multicolumn{2}{c}{\textbf{\ac{CoM} error }[\SI{}{\centi\meter}]} & \multicolumn{2}{c}{\textbf{Constraint violation }[\SI{}{\meter\per\second^2}]}  \\
& \ac{PR}     & \ac{LA}     & \ac{PR} & \ac{LA}    \\
\midrule
\SI{0}{\milli\second}    & 1.68            & 1.68      & 0.00          & 0.00     \\
\SI{3}{\milli\second}    & 4.52            & \textbf{1.68}      & 5.21           & \textbf{0.02}     \\
\SI{5}{\milli\second}    & $\infty$            & \textbf{1.71}      & $\infty$           & \textbf{0.03}     \\
\SI{25}{\milli\second}    & $\infty$            & \textbf{1.89}      & $\infty$           & \textbf{0.41}     \\
\bottomrule
\end{tabular}
\vspace{-0.6cm}
\end{table}

\begin{figure}[!t]
\centering

\subfloat[\ac{PR} control scheme]{\includegraphics[width=0.5\linewidth]{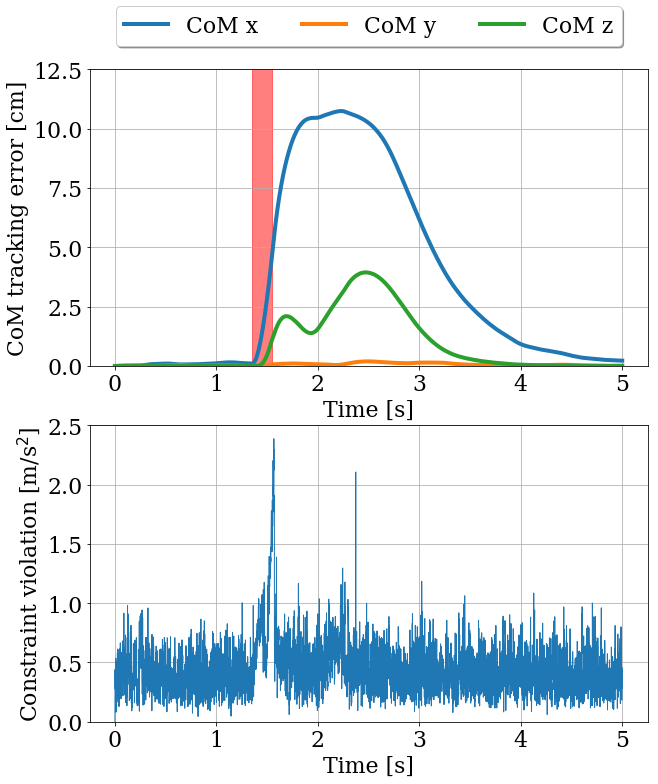}}
\subfloat[\ac{LA} control scheme]{
     \includegraphics[width=0.5\linewidth]{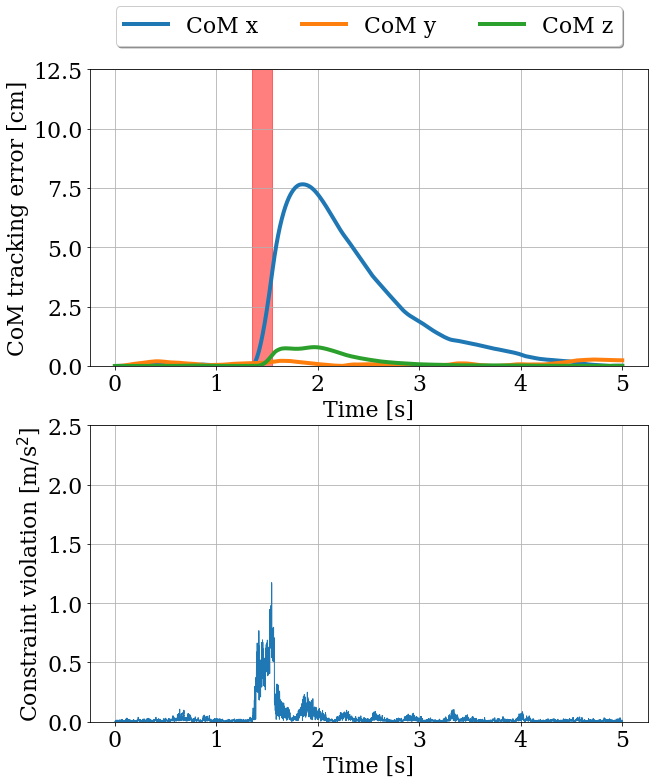}}
\caption{Typical control performance of \ac{PR} and \ac{LA} for the balancing task in the smart factory scenario. Red shaded area indicates the time window where we apply a push.}
\label{fig:balance}
\vspace{-0.3cm}
\end{figure}

\paragraph{Simulated delays with blockage}
As described in Sec.~\ref{sec:network_sim}, we simulated two different scenarios and generated $100$ delay profiles for each scenario. For the balancing task, both the naive remote controller \ac{PR} and \ac{LA} managed to keep the robot in balance with high success rate. However, even though \ac{PR} was able to complete the task, \ac{LA} was still advantageous in the sense that it reduced the \ac{CoM} tracking error and the constraint violation. As shown in Fig.~\ref{fig:balance}, the constraint violation was significantly reduced except when the robot was being pushed. 

For the walking task, we simulated on-board full \ac{QP} computation as described in Sec.~\ref{sec:onboard} by restricting the computation on a single core of the CPU at \SI{1.2}{\giga\hertz} within a time window of length \SI{100}{\milli\second} centered at the planned contact switch time. In addition, the pre-computation for contact switch was initiated \SI{10}{\milli\second} before the contact switch.
While real hardware implementation will be required to obtain a better estimate of the delay upper bound, \SI{10}{\milli\second} is a reasonable value as we will show later that the full \ac{QP} for the tasks can be solved in less than \SI{3}{\milli\second} on a single core of the CPU at \SI{1.2}{\giga\hertz}.
The experiments have shown that the naive approach \ac{PR} could not prevent the robot from falling on the ground in either of the scenarios, while \ac{LA} completed the walking task with a lower success rate in the smart factory scenario. This suggests that higher delay peaks, even with less frequent occurrence, is more damaging to the control performance than frequent delay peaks of smaller magnitude. This is particularly
relevant when there is a large change in the optimal active set, for example when changing contacts.

\begin{figure*}[!t]
\centering
\includegraphics[width=.9\textwidth]{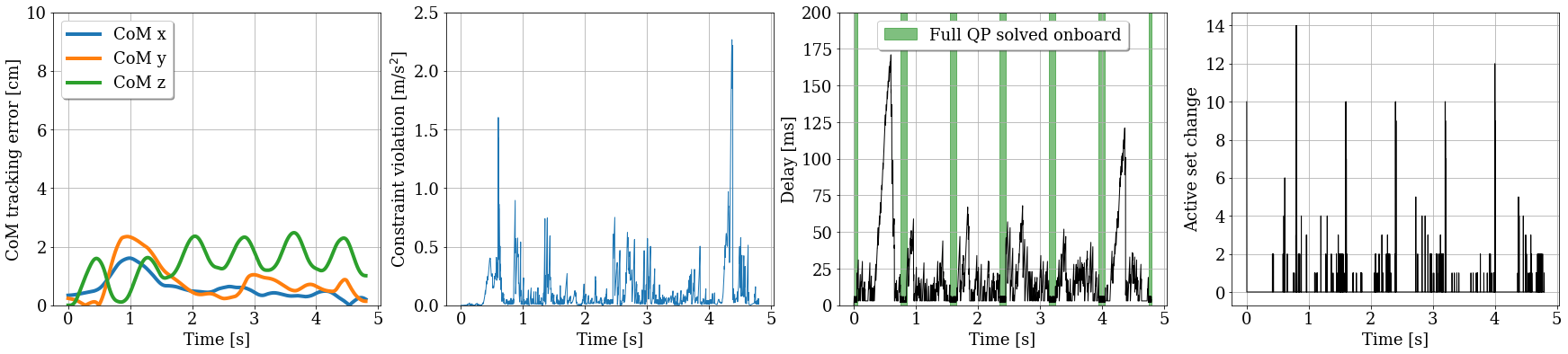}
\caption{Typical delay profile and control performance of \ac{LA} (walking in the smart factory scenario). Measurement noise is removed in the plot to better illustrate the relationship between delays and performance metrics.}
\label{fig:walk_local}
\vspace{-0.6cm}
\end{figure*}

Fig.~\ref{fig:walk_local} illustrates the control performance of \ac{LA} in one instance of the smart factory scenario. It can be seen that there is a correlation between higher delays and larger constraint violation; especially around \SI{0.5}{\second} and \SI{4.2}{\second} the two delay peaks  have caused very large constraint violation. On the other hand, while the pre-computation of the contact switch caused large discrepancy between the cached and the true optimal active set, the control performance was not significantly deteriorated due to the effective lower delay permitted by the on-board computation of full \acp{QP}.
\begin{table}[!t]
\centering
\caption{Task performance of \ac{PR} and \ac{LA} in the two scenarios. Performance metrics are computed from only successful trials.}
\label{tab:ns3_result}
\begin{tabular}{cccccc}
\toprule
\multirow{2}{*}{\textbf{Scenario}} & \multirow{2}{*}{\textbf{Metrics}} & \multicolumn{2}{c}{\ac{PR}} & \multicolumn{2}{c}{\ac{LA}}  \\
& & Balance     & Walk     & Balance    & Walk  \\
\midrule
\multirow{3}{*}{\textbf{Factory}} & \tiny Success rate & 98$\%$ & $0\%$ & $\mathbf{99\%}$ & $\mathbf{85\%}$ \\
& \tiny \ac{CoM} error [\SI{}{\centi\meter}] & 2.60 & - & \textbf{1.46} & \textbf{2.19} \\
& \tiny Constraint violation [\SI{}{\meter\per\second^2}] & 0.77 & - & \textbf{0.03} & \textbf{0.84} \\
\multirow{3}{*}{\textbf{Building}} & \tiny Success rate & $\mathbf{100\%}$ & $0\%$ & $\mathbf{100\%}$ & $\mathbf{100\%}$ \\
& \tiny \ac{CoM} error [\SI{}{\centi\meter}] & 1.39 & - & \textbf{1.37} & \textbf{1.83} \\
& \tiny Constraint violation [\SI{}{\meter\per\second^2}] & 0.40 & - & \textbf{0.03} & \textbf{0.05} \\
\bottomrule
\end{tabular}
\vspace{-0.6cm}
\end{table}

\begin{table}[!t]
\centering
\caption{Computation time in milliseconds of the local controller and the full \ac{QP} for the walking task on a single core of the CPU at \SI{1.2}{\giga\hertz} and \SI{3.7}{\giga\hertz} respectively.}
\label{tab:complexity}
\begin{tabular}{ccccc}
\toprule
\multirow{2}{*}{\textbf{CPU frequency}} & \multicolumn{2}{c}{\textbf{Average}} & \multicolumn{2}{c}{\textbf{Worst}}  \\
& \textbf{QP}     & \textbf{Local}     & \textbf{QP}    & \textbf{Local}    \\
\midrule
\SI{3.7}{\giga\hertz}            & 0.36            & \textbf{0.09}      & 0.70           & \textbf{0.22}     \\
\SI{1.2}{\giga\hertz}            & 1.21            & \textbf{0.30}      & 2.39           & \textbf{0.76} \\
\bottomrule
\end{tabular}
\vspace{-0.6cm}
\end{table}
\paragraph{Robustness to active set discrepancy}
Our approach assumes that the optimal active set does not change during network delays or communication loss. However, it is inevitable that the cached active set may differ from the true one when the delays are too high and the active set rapidly changes, for example, when the robot switches contact or receives an unexpected push. While we have shown that our approach enabled the robot to complete the task under such circumstances, we further demonstrate empirically how the discrepancy between the cached optimal active set and the true one, hence the cardinality $|(\mathcal{A}_{\text{opt}}\cup\mathcal{\hat{A}}_{\text{opt}})\backslash (\mathcal{A}_{\text{opt}}\cap\mathcal{\hat{A}}_{\text{opt}})|$, affects the control performance in Table~\ref{tab:activeset_walk}. Recall that all the equality constraints belong to the active set and the only inequality constraints we imposed are friction cone constraints, hence the change in the optimal active set is always caused by the contact force reaching the friction cone limit. For the walking task, the increased number of different constraints in the active set neither suggests higher \ac{CoM} tracking error nor higher constraint violation. Indeed, the control performance deteriorates the most when there are about 5 different active constraints. For the balancing task on the other hand, the control performance worsens as long as there is any different active constraints---here we note that the source of the discrepancy in the optimal active set is different for the two tasks considered: for the balancing task, the cached optimal active set differs from the true one when the external push was applied; in the walking task, discrepancy occurs when the robot approaches contact switch, most notably when the pre-computed solution for the next contact mode was used.

\begin{table}[!t]
\centering
\caption{Performance of \ac{LA} in the burning building scenario when the cached optimal active set differs from the true one. The cardinality of active set discrepancy is defined as $|(\mathcal{A}_{\text{opt}}\cup\mathcal{\hat{A}}_{\text{opt}})\backslash (\mathcal{A}_{\text{opt}}\cap\mathcal{\hat{A}}_{\text{opt}})|.$}
\label{tab:activeset_walk}
\begin{tabular}{ccccc}
\toprule
{\textbf{\scriptsize Cardinality of}} & \multicolumn{2}{c}{\textbf{\scriptsize \ac{CoM} error }[\SI{}{\centi\meter}]} & \multicolumn{2}{c}{\textbf{\scriptsize Constraint violation }[\SI{}{\meter\per\second^2}]}  \\
\textbf{\scriptsize active set discrepancy} & \emph{\scriptsize Walk}     & \emph{\scriptsize Balance}     & \emph{\scriptsize Walk} & \emph{\scriptsize Balance}    \\
\midrule
0    & 1.78            & 0.01      & 0.05           & 0.02     \\
1    & 1.91            & 0.05      & 0.06           & 0.19     \\
2    & 1.88            & 0.05      & 0.06           & 0.15    \\
3    & 1.90            & 0.06      & 0.08           & 0.13     \\
4    & 2.06            & 0.06      & 0.10           & 0.12     \\
5    & 2.06            & 0.06      & 0.19           & 0.14     \\
6    & 1.92            & 0.06      & 0.24           & 0.12     \\
7    & 1.54            & 0.06      & 0.09           & 0.12     \\
8    & 1.45            & 0.06      & 0.08           & 0.12     \\
9    & 1.38            & 0.06      & 0.06           & 0.11     \\
10   & 1.44            & 0.06      & 0.17           & 0.10     \\
11   & 1.40            & 0.06      & 0.18           & 0.10     \\
12   & 1.47            & -         & 0.19           & -     \\
13   & 1.52            & -         & 0.18           & -     \\
14   & 1.41            & -         & 0.19           & -     \\

\bottomrule
\end{tabular}
\vspace{-0.6cm}
\end{table}

\paragraph{Computation time}
We report in Table~\ref{tab:complexity} the average and worst-case computation time of
\begin{enumerate*}
    \item solving the full \ac{QP}~\eqref{eq:qp_invdyn}; and
    \item constructing the local controller~\eqref{eq:qp_sol}
\end{enumerate*} for the walking task in the burning building scenario. The computation time was obtained when the computation was restricted on a single core of the CPU at \SI{1.2}{\giga\hertz} and \SI{3.7}{\giga\hertz} respectively to emulate a low-power on-board computer required by \ac{LA} and a high-performance computer by traditional purely local optimization-based whole body inverse dynamics controller. It can be seen that the local controller computation takes significantly less time than solving a full \ac{QP}. This is not surprising, as the time complexity of the \ac{QP} is dominated by the \ac{HCOD} of cubic complexity $O(2n^3)$ where $n$ is the number of decision variables; in our case, $n = 61$. In the case of changes in the active-set, the decomposition is updated in approximately $O(2n^2)$ operations. On the other hand, the local controller computation is only dominated by a matrix-vector multiplication of time complexity $O(n^2)$ and can be parallelized if needed.




\section{Conclusion}
\label{sec:conclusion}
In this work, we have presented the first edge-based whole-body control algorithm over a \ac{5G} wireless link subject to unpredictable realistic delays. The proposed algorithm complements the remote controller on the network edge to robustly complete the task when there is a temporary unexpected communication delay. Simulation results have shown that the algorithm significantly improves control performance for balancing and walking tasks under both constant and stochastic delays. The proposed local controller is much more efficient than solving a full \ac{QP}, such that it can be executed on a low-power on-board computer with limited computational resources. We do emphasize that our current framework requires solving full \acp{QP} onboard when switching contacts. Although this is a benign requirement, it would be an interesting future research direction to explore the possibilities of handling contact switches when delay occurs, without solving the full \ac{QP} locally. For example, we can monitor and predict the channel quality online and only make contact switch when we are confident in the connectivity. Another option would be computing offline the \ac{QP} solution in an explicit MPC manner so that it can be stored locally on the robot and be inquired when needed.




\bibliographystyle{IEEEtran}
\bibliography{root}
\end{document}